\documentclass[10pt,twocolumn,letterpaper]{article}

\usepackage{iccv}
\usepackage{times}
\usepackage{epsfig}
\usepackage{graphicx}
\usepackage{amsmath}
\usepackage{amssymb}
\usepackage{booktabs} 
\usepackage{array} 
\usepackage{lipsum} 

\usepackage[breaklinks=true,bookmarks=false]{hyperref}

\iccvfinalcopy 

\def\iccvPaperID{****} 

\ificcvfinal\fi

\begin{document}

\title{ICCV23 Visual-Dialog Emotion Explanation Challenge: SEU\_309 Team Technical Report}

\author{Yixiao Yuan\thanks{These authors contributed equally to this work.}\\
Columbia University\\
{\tt\small yixiao.yuan@columbia.edu}
\and
Yingzhe Peng\footnotemark[1]\\
Southeast University\\
{\tt\small yingzhe.peng@seu.edu.cn}
}

\maketitle
\ificcvfinal\thispagestyle{empty}\fi

\begin{abstract}
The Visual-Dialog Based Emotion Explanation Generation Challenge focuses on generating emotion explanations through visual-dialog interactions in art discussions. Our approach combines state-of-the-art multi-modal models, including Language Model (LM) and Large Vision Language Model (LVLM), to achieve superior performance. By leveraging these models, we outperform existing benchmarks, securing the top rank in the ICCV23 Visual-Dialog Based Emotion Explanation Generation Challenge, which is part of the 5th Workshop On Closing The Loop Between Vision And Language (CLCV) with significant scores in F1 and BLEU metrics. Our method demonstrates exceptional ability in generating accurate emotion explanations, advancing our understanding of emotional impacts in art.
\end{abstract}

\section{Introduction}

The Visual-Dialog Based Emotion Explanation Generation Challenge~\cite{haydarov2023affective} focuses on the nuanced task of generating emotion explanations based on visual-dialog interactions within the context of art discussions. This unique challenge stems from the complexity of understanding emotional responses elicited by artworks, which often transcend straightforward visual cues to encompass historical, cultural, and subjective interpretations. Our team, SEU\_309, introduces an innovative approach that combines state-of-the-art multi-modal models. We train two independent models with Language Model (LM) and Large Vision Language Model (LVLM), and we combine the results together to achieve superior performance. In the LM-based method, we use the BLIP2~\cite{li2023blip} model to convert the image into text, and afterwards, we concatenate the generated text. In the LVLM-based method, we use the LLAVA~\cite{liu2023visual} to combine image and text input. It's a novel end-to-end trained large multimodal model that combines a vision encoder and Vicuna for general-purpose visual and language understanding.

Our method enhances the accuracy of emotion explanation generation, outperforming existing benchmarks across various metrics. The culmination of our efforts was demonstrated in our achieving the top rank in the Visual-Dialog Based Emotion Explanation Generation Challenge, with our team, SEU\_309, securing the first place with an F1 score of 52.36 and a BLEU score of 0.26, for a total score of 26.31. This achievement underscores our model's exceptional ability to accurately generate emotion explanations. The integration of visual and dialogic elements into our model not only addresses the challenge posed by the subjective nature of art appreciation but also aligns with the broader imperative for AI systems to be attuned to human emotions. 

In summary, by leveraging advanced multimodal models and a novel dataset, we are able to generate accurate and contextually rich emotion explanations. This not only advances our understanding of the emotional impact of art but also contributes to the broader goal of creating AI systems that are more aligned with human emotional states and needs.

\section{Related Work}

Multimodal learning in the intersection of visual and linguistic data has seen significant advancements in recent years. This area of research focuses on developing models that can understand and generate content by integrating information from both visual inputs, such as images or videos, and textual descriptions. These models leverage the complementary nature of visual and textual data to perform a wide range of tasks, including but not limited to image captioning, visual question answering, and cross-modal information retrieval.

One of the key innovations in this field has been the development of models that are trained on vast datasets of image-text pairs.~\cite{radford2021learning, li2023blip, li2022blip} These models employ contrastive learning techniques to align the representations of images and their corresponding textual descriptions in a shared embedding space. By doing so, they learn to associate specific visual features with relevant linguistic concepts, enabling them to understand complex queries and content with a high degree of accuracy. Moreover, the use of transformer-based~\cite{vaswani2017attention} architectures has further enhanced the capabilities of these models. Transformers provide a flexible and powerful framework for modeling sequential data, and their adoption in multimodal learning has allowed for the effective integration of visual and textual inputs. This has led to significant improvements in tasks requiring nuanced understanding and generation of content, such as generating descriptive captions for images that accurately reflect their content and context or answering questions based on visual information.

The emergence of Large Language Models (LLMs) has also marked a significant milestone in the field of artificial intelligence, particularly in natural language processing. Models like GPT ~\cite{achiam2023gpt}, BERT ~\cite{devlin2018bert}, and others have demonstrated exceptional abilities in generating and understanding text, significantly advancing the capabilities of AI in understanding human languages. These models have been instrumental in a variety of applications, including text generation, translation, and semantic analysis, showcasing their versatility and powerful computational abilities. However, their primary limitation lies in their text-only nature, confining their applicability to tasks that do not require understanding or generating content in other modalities such as images, speech, and videos.

To bridge this gap, the research and development of Large Vision Language Models (LVLMs)~\cite{liu2023improved, bai2023qwen, liu2023visual, zhu2023minigpt, ye2023mplugowl} have gained momentum. LVLMs are designed to perceive and understand both textual and visual information, thereby broadening the scope of applications for AI models. This integration allows for a more holistic understanding of content, enabling models to perform tasks that involve both text and images, such as image captioning, visual question answering, and visual grounding. 

LVLMs are built upon the foundation of LLMs, incorporating additional components that allow them to process and understand visual data. These components typically include visual receptors for image processing, input-output interfaces that enable the model to handle multimodal data, and training pipelines that are tailored to accommodate the complexities of learning from both text and images. Furthermore, these models are often trained on multilingual multimodal cleaned corpora, enhancing their ability to function across different languages and modalities.

The development of LVLMs represents a significant step forward in the field of AI, offering new possibilities for applications that require an understanding of the world that extends beyond text. By combining the capabilities of LLMs with the ability to process visual information, LVLMs are paving the way for more advanced and versatile AI systems. These models have set new benchmarks in various visual-centric tasks, demonstrating their potential in enhancing the performance and applicability of AI across a broad range of domains.

\section{Method}
We use two methods to incorporate image information: one is to directly convert the image into a caption using an Image Captioning model and input it into the language model, and the second is to utilize state-of-the-art LVLMs by directly inputting the image.
\subsection{LM-Based}
We follow~\cite{haydarov2023affective} to use the BLIP2~\cite{li2023blip} model to convert the image into text. Afterwards, we concatenate the generated text according to the instructions provided in Table~\ref{tab:tab1}. Then, we fine-tune a pretrained language model. During the loss computation, we only calculate the loss for the Response part. Additionally, we are aware that The Visual-Dialog Based Emotion Explanation Generation Challenge requires both emotion classification and explanation generation. For the emotion classification part, we can utilize model ensemble to reduce classification bias. Specifically, we divide the dataset into 5 folds and train a language model for each fold. During the final prediction, we use the voting method for emotion classification.

\subsection{LVLM-Based}
\begin{figure}[htbp]
\centering
\includegraphics[width=\linewidth]{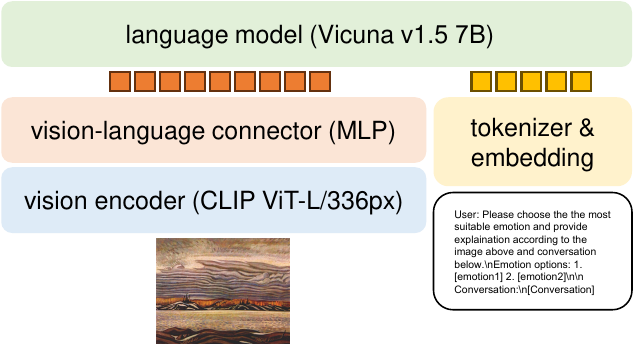}
\caption{LVLM-based method architecture}
\label{fig:yourlabel}
\end{figure}

To utlize the development of LVLM, we use instruct finetuning to train LVLM model to generate emotion and explaination. We follow~\cite{liu2023visual} to combine image and text input. It's a novel end-to-end trained large multimodal model that combines a vision encoder and Vicuna for general-purpose visual and language understanding. The architecture is provided in~\ref{fig:yourlabel}. We concatenate the generated text according to the instructions provided in Table~\ref{tab:tab2}.

\begin{table*}[!h]
\centering
\caption{The instruction template of LM-based experiments.}
\label{tab:tab1}
\resizebox{0.8\textwidth}{!}{%
\begin{tabular}{@{}ll@{}}
\toprule
\textbf{Input} & \parbox[t]{10cm}{%
    \small\textless{}emotion\textgreater{}[emotion1]\small\textless{}emotion\textgreater{}[emotion2]\small\textless{}caption\textgreater{}[BLIP2Caption]\\\small\textless{}conversation\textgreater{} [Conversation] I feel%
} \\

\midrule
\textbf{Response} & \parbox[t]{10cm}{%
    \small[Label Emotion] because [Explanation]%
} \\
\bottomrule
\end{tabular}

}
\end{table*}

\begin{table*}[!h]
\centering
\caption{The instruction template of LVLM-based experiments.}
\label{tab:tab2}
\resizebox{0.8\textwidth}{!}{%
\begin{tabular}{@{}ll@{}}
\toprule
\textbf{Input} & \parbox[t]{10cm}{%
    \small<image>Please choose the most suitable emotion and provide explanation according to the image above and conversation below.\\Emotion options: 1. [emotion1] 2. [emotion2]\\\\
Conversation:\\{[}Conversation{]}
} \\
\midrule
\textbf{Response} & \parbox[t]{10cm}{%
    \small Choice:{[}Label Emotion{]}\\Explanation: {[}Explanation{]}%
} \\
\bottomrule
\end{tabular}
}
\end{table*}
\section{Experimental Setup} 
In the LM-based experiments, we use Bart-Large~\cite{lewis2019bart} as our language model. We set the batch size to 32 and the learning rate to 5e-5 with 3 epochs. We accumulate gradients for two steps of mini-batches before performing backpropagation. We use the AdamW~\cite{loshchilov2017decoupled} optimizer and a Cosine learning rate scheduler. Additionally, we employ the warmup technique, where we gradually increase the learning rate during the first 1\% of steps. All experiments are conducted using mixed precision training and trained on a single A6000 GPU. We employ cross-validation by randomly dividing the training set into 5 folds.

In the LVLM-based experiment, we incorporate Low-Rank Adaptation (LoRA) to enhance parameter tuning efficiency, employing a rank configuration of 96 and an alpha parameter of 192. The initial learning rate is set to \(5 \times 10^{-5}\). Preliminary experiments with a learning rate of \(2 \times 10^{-4}\) resulted in a notable loss spike, a phenomenon commonly encountered during the training of Large Language Models (LLMs) ~\cite{mlEngineeringInstabilities}. For our foundational model, we utilize `llava-v1.5-7b`, augmented with `clip-vit-large` as the vision module. The training regimen consists of a batch size of 16 over two epochs. All experiments are
conducted and trained on a single RTX3090Ti GPU.

\section{Results and Analysis}
\subsection{LM-based}
Table~\ref{tab:performance_scores} show the average of weighted-f1 and the bleu scores for our 5-Fold model in validation set. We can observe that models trained on different folds have their own merits in terms of Weighted F1 and BLEU metrics. Specifically, The Fold 5 can get best Weighted F1 while for the Fold 4 get the best BLEU scores. We note that for emotion classification, we can use voting method to ensemble all model's predictions, while for explanation generation, we can not to ensemble these models because the autoregressive manner. Therefore, for emotion explanation, we choose the best model (Fold 4) in BLEU performance.

Tables~\ref{tab:model_comparison} show the CV performance and the performance in leaderboard. For BLEU in ensemble model, we use the explanation generated by Fold 4. We find that ensembling can greatly improve the performance of emotion classification. On the validation set, the Ensemble model show an improvement of 0.827 in weighted F1 compared to a single model (Fold 5). On the leaderboard, the Ensemble model exhibited a weighted F1 improvement of 0.935 compared to Fold 5. This demonstrates that ensembling enhances the generalizability of the model in emotion classification task. 

\begin{table}[!h]
\centering
\caption{Performance Scores Across Folds in LM-based.}
\label{tab:performance_scores}
\begin{tabular}{@{}lcc@{}}
\toprule
& \textbf{Weighted F1} & \textbf{BLEU} \\
\midrule
Fold 1 & 51.058 & 0.2416 \\
Fold 2 & 51.536 & 0.2394 \\
Fold 3 & 51.869 & 0.2390 \\
Fold 4 & 51.942 & 0.2429 \\
Fold 5 & 52.368 & 0.2402 \\
\midrule
Average & 51.755 & 0.2406 \\
\bottomrule
\end{tabular}
\end{table}

\begin{table}[ht]
\centering
\caption{Comparison of Ensemble Model and Fold 5 Scores, where Val means test in validation set and LB means test in leaderboard. }
\label{tab:model_comparison}
\begin{tabular}{@{}lcc@{}}
\toprule
& \textbf{Weighted F1} & \textbf{BLEU} \\
\midrule
Ensemble Model (Val) & 52.363 & 0.2429 \\
Fold 5 (Val) & 51.536 & 0.2402 \\
\midrule
\addlinespace 
Ensemble Model (LB) & 52.016 & 0.2327 \\
Fold 5 (LB) & 51.425 & 0.2390 \\
\bottomrule
\end{tabular}
\end{table}

\subsection{LVLM-based}
Table~\ref{tab:lvlm} presents our leaderboard performance. Owing to computational resource constraints, we were unable to employ k-fold validation on the LVLM-based method.
\begin{table}[ht]
\centering
\caption{LVLM-based Method Performance Score}
\label{tab:lvlm}
\begin{tabular}{@{}lcc@{}}
\toprule
& \textbf{Weighted F1} & \textbf{BLEU} \\
\midrule
LVLM-based (LB) & 48.371 & 0.2641 \\
\bottomrule
\end{tabular}
\end{table}
\subsection{Combining LM-based and LVLM-based Methods}
Our comparative analysis indicates that the LM-based models excel in emotion classification, as evidenced by their superior Weighted F1 scores, while LVLM-based models show a notable advantage in explanation generation, demonstrated by higher BLEU scores. To leverage the strengths of both approaches, we propose a hybrid strategy. Specifically, we employ a hard voting mechanism using the emotion classification results from both the LM-based and LVLM-based models to determine the final emotion classification. For the explanation generation, we directly utilize the explanations provided by the LVLM-based model. This combination strategy aims to optimize overall performance by integrating the precise emotion classification capabilities of the LM-based models with the nuanced explanation generation of the LVLM-based models. Consequently, this hybrid approach achieves a final performance with a Weighted F1 score of 52.361 and a BLEU score of 0.2641, effectively harnessing the complementary strengths of both methodologies.

\section{Conclusion}
Our research contributes significantly to the field of emotion explanation generation by combining LM and LVLM approaches. We demonstrate that a hybrid strategy leveraging the strengths of both models enhances overall performance. The LM-based models excel in emotion classification, while LVLM-based models provide nuanced explanation generation. This combination yields a robust solution with high accuracy in both emotion classification and explanation generation, highlighting the potential of multi-modal approaches in understanding and interpreting complex emotional responses in art.
{\small
\bibliographystyle{ieee_fullname}
\bibliography{egbib}
}

\end{document}